\newcommand{\trflag}{1}
\documentclass[letterpaper]{article} \usepackage{aaai24}  \usepackage{times}  \usepackage{helvet}  \usepackage{courier}  \usepackage[hyphens]{url}  \usepackage{graphicx}    \usepackage{natbib}  \usepackage{caption}       \usepackage{algorithm}
\usepackage{algorithmic}

\usepackage{newfloat}
\usepackage{listings}
\DeclareCaptionStyle{ruled}{labelfont=normalfont,labelsep=colon,strut=off} 
\floatstyle{ruled}
\newfloat{listing}{tb}{lst}{}
\floatname{listing}{Listing}
\usepackage{tikz}
\usetikzlibrary{automata,calc,arrows}

\usepackage{xspace}

\usepackage{array}
\usepackage{subcaption}
\usepackage{mathtools}
\usepackage{amsfonts}
\usepackage{booktabs}
\usepackage{multirow}
\usepackage{amsmath}
\usepackage{stmaryrd}
\usepackage[noabbrev]{cleveref}
\usepackage{fixmath}
\usepackage{amsthm}
\usepackage{csquotes}
\usepackage{braket}
\usepackage{todonotes}
\newtheorem{definition}{Definition}
\newtheorem{theorem}{Theorem}

\newtheorem{corrolary}{Corrolary}
\newtheorem{proposition}{Proposition}

\newcommand{\var}[1]{{\operatorname{\mathit{#1}}}}

\title{On the Computational Complexity of Stackelberg Planning and Meta-Operator Verification\iftr{: Technical Report}}
\author {
Gregor Behnke\textsuperscript{\rm 1},
    Marcel Steinmetz\textsuperscript{\rm 2}
}
\affiliations {
\textsuperscript{\rm 1}ILLC, University of Amsterdam, Amsterdam, The Netherlands\\
    \textsuperscript{\rm 2}LAAS-CNRS, Universit\'{e} de Toulouse, France\\
    g.behnke@uva.nl, marcel.steinmetz@laas.fr
}

\newcommand{\iftr}[1]{\ifthenelse{\isundefined{\trflag}}{}{#1}}

\newcommand{\ifnottr}[1]{\ifthenelse{\isundefined{\trflag}}{#1}{}}

\newcommand{\ie}{i.e.}

\newcommand{\defined}[1]{\emph{#1}}

\newcommand{\naturals}{\mathbb{N}}

\newcommand{\bigO}{\mathcal{O}}

\newcommand{\tm}{\mathit{TM}\xspace}
\newcommand{\poly}{P\xspace}
\newcommand{\np}{\textrm{NP}\xspace}

\newcommand{\conpspace}{\textrm{coNPSPACE}\xspace}

\newcommand{\npspace}{\textrm{NPSPACE}\xspace}
\newcommand{\pspace}{\textrm{PSPACE}\xspace}

\newcommand{\sigmaP}[1]{\ensuremath{\Sigma^{\textrm{P}}_{#1}}\xspace}
\newcommand{\piP}[1]{\ensuremath{\Pi^{\textrm{P}}_{#1}}\xspace}
\newcommand{\sigmaSecond}{\sigmaP{2}}
\newcommand{\piSecond}{\piP{2}}

\newcommand{\planunsat}{\textrm{PLANUNSAT}\xspace}
\newcommand{\plansat}{\textrm{PLANSAT}\xspace}
\newcommand{\planopt}{\textrm{PLANMIN}\xspace}
\newcommand{\stackelsat}{\textrm{STACKELSAT}\xspace}
\newcommand{\stackelopt}{\textrm{STACKELMIN}\xspace}
\newcommand{\stackelpoly}{\textrm{STACKELPOLY}\xspace}
\newcommand{\metaverif}{\textrm{METAOPVER}\xspace}
\newcommand{\polymetaverif}{polyMETAOPVER\xspace}

\newcommand{\tuple}[1]{\langle #1\rangle}

\newcommand{\task}{\Pi}
\newcommand{\facts}{V}
\newcommand{\actions}{A}
\newcommand{\init}{I}
\newcommand{\goal}{G}
\newcommand{\pre}{\mathop{pre}}

\newcommand{\add}{\mathop{add}}
\newcommand{\del}{\mathop{del}}
\newcommand{\cost}{c}
\newcommand{\apply}[1]{\llbracket #1 \rrbracket}
\newcommand{\plan}{\pi}
\newcommand{\bound}{B}
\newcommand{\meta}{\sigma}

\newcommand{\leader}[1]{#1^{L}}
\newcommand{\follower}[1]{#1^{F}}
\newcommand{\stask}{\task^{LF}}
\newcommand{\ftask}[1]{\task^{F}(#1)}
\newcommand{\lactions}{\leader{\actions}}
\newcommand{\factions}{\follower{\actions}}
\newcommand{\fgoal}{\follower{\goal}}
\newcommand{\lplan}{\leader{\plan}}
\newcommand{\fplan}{\follower{\plan}}
\newcommand{\fcost}{\follower{\cost}}
\newcommand{\lbound}{\leader{\bound}}
\newcommand{\fbound}{\follower{\bound}}
\newcommand{\dominates}{\sqsubset}
\newcommand{\wdominates}{\sqsubseteq}

\newif\iflong

\iftr{\nocopyright}

\begin{document}

\maketitle

\begin{abstract}
Stackelberg planning is a recently introduced single-turn two-player
  adversarial planning model, where two players are acting in a joint
  classical planning task, the objective of the first player being hampering the
  second player from achieving its goal. 
This places the Stackelberg planning problem somewhere between
  classical planning and general combinatorial two-player games. But,
  where exactly?
All investigations of Stackelberg planning so far focused on practical
  aspects.
We close this gap by conducting the first theoretical complexity analysis of
  Stackelberg planning. We show that in general Stackelberg planning is actually
  no harder than classical planning. Under a polynomial plan-length restriction,
  however, Stackelberg planning is a level higher up in the polynomial
  complexity hierarchy, suggesting that compilations into classical planning
  come with a worst-case exponential plan-length increase. In attempts to identify
  tractable fragments,
we further study its complexity under various planning task
  restrictions, showing that Stackelberg planning remains intractable where
  classical planning is not.
We finally inspect the complexity of meta-operator verification, a problem that has been recently connected to Stackelberg planning.

\end{abstract}

\section{Introduction}

  Stackelberg planning~\cite{speicher2018stackelberg} is an adversarial planning
  problem, in which two agents/players act consecutively in a joint classical
  planning task. The objective of the first player (called the \emph{leader}) is
  to choose and to play a plan that maximally raises the cost of the second
  player (the \emph{follower}) to subsequently achieve its goal. This type of
  planning is useful for real-world adversarial settings commonly found in the
  cyber-security domain~\cite{speicher2018formally,di2023pareto}.
To solve Stackelberg planning tasks, there so far exists just a single generic
  algorithm paradigm called \emph{leader-follower search}
  \cite{speicher2018stackelberg}, which searches over possible leader plans,
  solving a classical planning task for each. As the number of possible plans is
  exponential in the worst case, this makes one wonder how the complexity of
  Stackelberg planning relates to classical planning. Focusing on algorithmic
  improvements
  \cite{speicher2018stackelberg,torralba2021faster,sauer2023lifted}, existing
  works have neglected this question so far.

  We close this gap, providing the first theoretical analysis of Stackelberg
  planning's complexity. Stackelberg planning is a special case of general
  combinatorial two-player games \cite{StockmeyerC79}. And, indeed, as many
  combinatorial games can, Stackelberg planning is reducible to fully-observable
  non-deterministic (FOND) planning \cite{CimattiRT98}, using action effect
  non-determinism to emulate all possible follower choices. This narrows down
  Stackelberg planning's complexity to the range between classical planning
  \cite{Bylander1994} and FOND planning \cite{Littman97}.
We show that Stackelberg planning is \pspace-complete, and thus is in fact not
  harder than classical planning in general. However, Stackelberg planning is
  $\sigmaSecond$-complete under a polynomial plan-length restriction. This
  relates to results in FOND \cite{rintanen1999constructing} and conformant
  planning \cite{baral2000computational}, and contrasts the $\np$-completeness
  of the corresponding classical planning problem~\cite{JonssonB98}. Hence,
  unless $\np = \sigmaSecond$, polynomial compilations of Stackelberg planning
  into classical planning have a worst-case exponential plan-length blow-up.

The analysis of tractable fragments has shown to be an important source for the
development of domain-independent heuristic in classical
planning~\cite[e.g.,][]{Hoffmann2001FF,Domshlak2015RedBlack}. With the vision of
establishing a basis for the development of leader-follower search heuristics,
we analyze the complexity of Stackelberg planning under
various syntactic restrictions. An overview of our results is given in
Tab.~\ref{table:overview}.

\begin{table*}[t]
\centering
\small
\newcommand{\tRef}[1]{ {\footnotesize (Theorem~#1)}}
\newcommand{\cRef}[1]{ {\footnotesize (Corollary~#1)}}
\newcommand{\bCite}{}

\begin{tabular}{>{\flushleft\arraybackslash}m{3cm}>{\centering\arraybackslash}m{.5cm}>{\centering\arraybackslash}m{4.5cm}>{\centering\arraybackslash}m{0.9cm}>{\centering\arraybackslash}m{3.2cm}>{\centering\arraybackslash}m{3.0cm}}
\toprule
& \multicolumn{2}{c}{Plan existence} & \multicolumn{2}{c}{Optimal planning} & \\
\cmidrule(lr){2-3} \cmidrule(lr){4-5}
Syntactic restrictions & \plansat & \stackelsat & \planopt & \stackelopt & \multirow{-2}{*}{\metaverif} \\
\midrule
{ $\ast$ preconds $\ast$ effects \newline $|\pi|$ not bounded } & \pspace  \bCite & { \pspace  \tRef{\ref{stackel:base}} } & \pspace  \bCite & \pspace  \tRef{\ref{thm:stackelbase-opt}}& \pspace  \tRef{\ref{thm:metaverif-base}} \\
\hline
{ $\ast$ preconds $\ast$ effects \newline $|\pi| \in \bigO(n^k)$} & \np  \bCite & { \sigmaSecond  \tRef{\ref{stackel:poly}} } & \np \bCite  & \sigmaSecond  \tRef{\ref{stackel:poly}}& \piSecond  \tRef{\ref{thm:metaverif-poly}} \\
\hline
{ 1 precond 1$+$ effect} & \np  \bCite & { \sigmaSecond  \tRef{\ref{thm:stackel-sat-1-1}} } & \np \bCite & \sigmaSecond  \cRef{\ref{cor:stackel-opt-1-1}}& -- \\
\hline
{ $\ast$$+$ preconds 1 effect} & \poly \bCite  & { \np \tRef{\ref{thm:stackel-sat-+-1}} } & \np \bCite & \sigmaSecond \tRef{\ref{thm:stackel-opt-+-1}}& --  \\
\hline
{ 0 preconds 2 effects} & \poly & { \poly for $\infty$ effects \tRef{\ref{thm:stackel-sat-0}} }  & \np \bCite & \sigmaSecond  \tRef{\ref{thm:stackel-opt-0-2}}& -- \\
\hline
{ 0 preconds 1 effect \newline non-unit cost} & \poly & { \poly for $\infty$ effects \tRef{\ref{thm:stackel-sat-0}} } & \poly & \np \tRef{\ref{thm:stackel-opt-0-1}}& -- \\

\bottomrule
\end{tabular}
\caption{Overview of our complexity results. For comparison, the \plansat and
\planopt columns show the complexity of classical planning under the respective
task restrictions, as given by \cite{Bylander1994}. All results prove
completeness with respect to the different complexity classes.
$\ast$ means arbitrary number, $+$ only positive, $\ast+$ arbitrary positive, and $n+$ $n$ positive.
}
\label{table:overview}
\end{table*}

Lastly, we explore a problem related to Stackelberg planning:
\emph{meta-operator}~\cite{Pham2023MetaOperators} verification.
Meta-operators are action-sequence wild cards, which can be instantiated freely
for every state satisfying the operator's precondition as long as operator's
effects match.
\citeauthor{Pham2023MetaOperators} have cast verifying whether a given action is a valid meta-operator as a Stackelberg
planning task.
We show that meta-operator verification PSPACE-complete and $\piSecond$-complete under a
polynomial plan-length restriction.

\ifnottr{Proofs are provided in a technical report \cite{Behnke24tr}.}

\section{Background}

\paragraph{Classical Planning}

We assume STRIPS notation~\cite{Fikes1971STRIPS}. A planning task is a
tuple $\task = \tuple{\facts, \actions, \init, \goal}$ consisting of a set of
propositional \defined{state variables} (or \defined{facts}) $\facts$, a set of
\defined{actions} $\actions$, an \defined{initial state} $\init \subseteq
\facts$, and a \defined{goal} $\goal \subseteq \facts$.
For $p \in \facts$, $p$ and $\neg p$ are called
\defined{literals}. A \defined{state} $s$ is a subset of $\facts$, with the
interpretation that all state variables not in $s$ do not hold in $s$.
Each action $a \in \actions$ has a \defined{precondition}
$\pre(a)$, a conjunction of literals, an \defined{add effect} (also called
positive effect) $\add(a) \subseteq \facts$, a \defined{delete effect} (negative
effect) $\del(a) \subseteq V$, and a non-negative \defined{cost} $\cost(a) \in
\naturals_0$.
A planning task has \defined{unit costs} iff for all actions $c(a) = 1$.
$a$ is applicable in a state $s$ iff $s \models \pre(a)$. Executing
$a$ in $s$ yields the state $s\apply{a} = (s \setminus
\del(a)) \cup \add(a)$. These definitions are extended to action sequences
$\plan$ in an iterative manner.
The cost of $\plan$ is the sum of costs of its actions.
$\plan$ is called an \defined{$s$-plan} if $\plan$ is applicable in $s$ and
$\goal \subseteq s\apply{\plan}$. $\plan$ is an \defined{optimal} $s$-plan if
$\cost(\plan)$ is minimal among all $s$-plans. An (optimal) plan for $\task$ is
an (optimal) $\init$-plan. If there is no $\init$-plan, we say that $\task$ is
\defined{unsolvable}. Two decision problem formulations of classical planning
are considered in the literature. \defined{\plansat} is the problem of given a
planning task $\task$, deciding whether there exists any plan for $\task$.
\defined{\planopt} asks, given in addition a (binary-encoded) cost bound
$\bound$, whether there is a plan $\plan$ for $\task$ with cost $\cost(\plan)
\leq \bound$. Both problems are known to be \pspace-complete
\cite{Bylander1994}.

\paragraph{Stackelberg Planning}

A Stackelberg planning task~\cite{speicher2018stackelberg} is a tuple $\stask =
\tuple{\facts, \lactions, \factions, \init, \fgoal}$, where the set of actions
is partitioned into one for each player. A \defined{leader plan} is an action
sequence $\lplan = \tuple{a^L_1, \dots, a^L_n} \in (\lactions)^n$ that is
applicable in $\init$. $\lplan$ induces the \defined{follower task}
$\ftask{\lplan} = \tuple{\facts, \factions, \init\apply{\lplan}, \fgoal}$. An
(optimal) \defined{follower response} to $\lplan$ is an (optimal) plan for
$\ftask{\lplan}$. We denote by $\fcost(\lplan)$ the cost of the optimal follower
response to $\lplan$, defining $\fcost(\lplan) = \infty$ if $\ftask{\lplan}$ is
unsolvable.
Leader plans are compared via a dominance order between cost pairs where
$\tuple{\cost^L_1, \cost^F_1}$ \defined{weakly dominates} $\tuple{\cost^L_2,
\cost^F_2}$ ($\tuple{\cost^L_1, \cost^F_1} \wdominates \tuple{\cost^L_2,
\cost^F_2}$), if $\cost^L_1 \leq \cost^L_2$ and $\cost^F_1 \geq \cost^F_2$.
$\tuple{\cost^L_1, \cost^F_1}$ (strictly) \defined{dominates} $\tuple{\cost^L_2,
\cost^F_2}$ ($\tuple{\cost^L_1, \cost^F_1} \dominates \tuple{\cost^L_2,
\cost^F_2}$), if  $\tuple{\cost^L_1, \cost^F_1} \wdominates \tuple{\cost^L_2,
\cost^F_2}$ and $\tuple{\cost^L_1, \cost^F_1} \neq \tuple{\cost^L_2,
\cost^F_2}$. To simplify notation, we write $\lplan_1 \dominates \lplan_2$ if
$\tuple{\cost(\lplan_1), \fcost(\lplan_1)} \dominates \tuple{\cost(\lplan_2),
\fcost(\lplan_2)}$. A leader plan $\lplan$ is optimal if it is not dominated by
any leader plan.
Previous works have considered algorithms for computing the set of all optimal
solutions, called the \defined{Pareto frontier}.

\section{Stackelberg Planning Decision Problems}

We distinguish between two decision-theoretic formulations of Stackelberg
planning, akin to classical planning:

\begin{definition}[\stackelsat]
Given $\stask$, \stackelsat is the problem of deciding whether there is a
leader plan $\lplan$ that makes $\ftask{\lplan}$ unsolvable.
\end{definition}

\begin{definition}[\stackelopt]
Given $\stask$, and two binary-encoded numbers $\lbound, \fbound \in \mathbb
N_0$. \stackelopt is the problem of deciding whether there is a leader plan
$\lplan$ with $\tuple{\cost(\lplan), \fcost(\lplan)} \wdominates
\tuple{\lbound, \fbound}$.
\end{definition}

Interpreting the leader's objective as rendering the follower's
objective infeasible, the first definition directly mirrors the \plansat
plan-existence decision problem. Similarly, the second definition mirrors
\planopt in looking for solutions matching a given quantitative cost bound.
It is worth mentioning that both decision problems are implicitly looking for
only a single point in the Pareto frontier, whereas previous practical works dealt with
algorithms computing this frontier entirely. In terms of computational
complexity, this difference is however unimportant. In particular, answering
even just a single \stackelopt question does in fact subsume the computation of
the entire Pareto frontier -- if the answer is no, one necessarily had to
compare the given bounds to \emph{every} element in the Pareto frontier.

As in classical planning, \stackelsat can be easily (with polynomial overhead)
reduced to \stackelopt:

\begin{proposition}
\stackelsat is polynomially reducible to \stackelopt.
\label{prop:sat-opt-reducible}
\end{proposition}

 \iftr\begin{proof}
Let $\stask = \tuple{\facts, \lactions, \factions, \init, \fgoal}$ be a
Stackelberg task. There is a leader plan satisfying \stackelsat iff there is a leader plan satisfying \stackelopt with 
$\lbound = 2^{|\facts|} \cdot \max_{a^L \in \lactions} \cost(a^L) $ and $\fbound
= 2^{|\facts|} \cdot \max_{a^F \in \lactions} \cost(a^F)$. Clearly, both bounds
can be computed in time linear in the size of $\stask$.
\end{proof}

Given that Stackelberg planning is a proper generalization of classical
planning, the Stackelberg decision problems are guaranteed to be at
least as hard as the respective classical planning decision
problem.
By applying Immerman–Szelepcsényi theorem~\cite{szelep1987method,immerman1988nondeterministic}, we can prove that it is also no
harder than classical planning in the general case:

\begin{theorem} \label{stackel:base}
\stackelsat is \pspace-complete.
\end{theorem}

 \iftr{\begin{proof}
  \underline{Membership:} \citet{savitch1970relationships} showed that $\pspace
  = \npspace$. It hence suffices to show that \stackelsat is contained in
  \npspace. Deciding plan-existence for a classical planning task is
  \pspace-complete \cite{Bylander1994}, so by Savitch's theorem, it is therefore
  also \npspace-complete. This makes the dual problem, \planunsat, \ie,
  determining whether a classical planning is unsolvable, a member of
  \conpspace. The Immerman–Szelepcsényi
  theorem~\cite{szelep1987method,immerman1988nondeterministic} proves that
  $\conpspace = \npspace$, and therefore by Savitch's theorem, $\conpspace =
  \pspace$. In other words, \planunsat is in \pspace, so there must be a deterministic polynomially space bounded
  Turing machine $\tm$ that accepts a classical planning task iff that task is
  unsolvable. We compose $\tm$ into a non-deterministic polynomially space
  bounded algorithm deciding \stackelsat. We can non-deterministically guess a
  leader plan and compute the resulting state $s^L$. We then use $\tm$ to check
  whether the follower's task $\tuple{\facts, \factions, s^L, \fgoal}$ is
  unsolvable. If it is, we return true, otherwise we return false.

\underline{Hardness:} We reduce from \planunsat, which, as we have just seen, is \pspace-complete. Given a classical planning
problem $\task=(\facts,\actions,\init,\goal)$, we create the Stackelberg
planning problem $\stask = (\facts,\emptyset,\actions,\init,\goal)$, i.e., we
treat all actions as follower actions. Clearly, there is a leader plan satisfying \stackelsat iff the classical planning task is unsolvable. 
\end{proof} 

 }

\begin{theorem}
\label{thm:stackelbase-opt}
\stackelopt is \pspace-complete.
\end{theorem}

 \iftr{\begin{proof}
  The idea is the same as in the proof of Theorem~\ref{stackel:base}.
  \underline{Membership:} Since deciding bounded-cost plan-existence is
  \pspace-complete \cite{Bylander1994}, the combination of Savitch's theorem and
  the Immerman–Szelepcsényi
  theorem~\cite{szelep1987method,immerman1988nondeterministic} imply
  that the dual problem, \ie, deciding that all plans have a cost of more than
  $c$ is also \pspace-complete. Let $\tm$ denote a poly-space Turing machine
  that solves that problem. Non-deterministically guess an applicable
  sequence of actions with cost at most $\lbound$ and compute the resulting state
  $s^L$. Apply $\tm$ to determine whether the classical planning task
  $\Pi=(V,A^F,s^L,G)$ is unsolvable for a plan-cost bound of $\fbound$. If it is,
  return true. If not, return false.
\\
  \underline{Hardness:} We reduce again from \planunsat (cf. proof of
  Theorem~\ref{stackel:base}). Given a classical planning task
  $\Pi=(V,A,I,G)$, we create the Stackelberg planning task $\Pi =
  (V,\emptyset,A,I,G)$, i.e., we treat all actions as follower actions. We set
  $B^L = 0$ and $B^F = 1 + 2^{|V|} \cdot \max_{a \in A} c(a)$. As in most of
  the following proofs, we set $B^F$ one unit higher than the maximum cost of
  non-redundant plan, ensuring that there is a leader plan satisfying \stackelopt
  if and only if $\Pi$ is unsolvable\footnote{Setting $B^F = 2^{|V|} \cdot \max_{a \in A} c(a)$ is not sufficient as the shortest follower plan can have that cost.}.
\end{proof}

 }

In spite of these results, algorithms for Stackelberg planning are significantly
more complicated than their classical planning counterparts. In particular, the
results raise the question of whether it is possible to leverage 
the classical planning methods directly for solving Stackelberg tasks via compilation. Polynomial
compilations necessarily exist as per the theorems, yet, it is interesting to
investigate which ``side-effects'' these might need to have. 
In order to investigate these questions, we turn to a more fine granular
analysis by considering the complexity under various previously studied
syntactic classes of planning tasks.

\section{Stackelberg Planning under Restrictions}

\subsection{Polynomial Plan Length}

For classical planning, it is commonly known that restricting the length of the
plans to be \emph{polynomial} in the size of the planning task description,
makes the decision problems become \np-complete \cite{JonssonB98}. 

\begin{definition}[Polynomial Stackelberg Decision]
Given $\stask$ with non-$0$ action costs, and two binary-encoded numbers $\lbound, \fbound \in \mathbb
N_0$ that are bounded by some polynomial $p \in \mathcal O(\ell^k)$ for $\ell = |\facts| + |\lactions| + |\factions|$.
\stackelpoly is the problem of deciding whether there is a
leader plan $\lplan$ such that $\tuple{\cost(\lplan), \fcost(\lplan)}
\wdominates \tuple{\lbound, \fbound}$.
\end{definition}

We restrict the action cost to be strictly positive, ensuring
that considering leader and follower plans with polynomial length is sufficient
to answer the decision problem.
\stackelpoly is harder than the corresponding classical problem. 

\begin{theorem}\label{stackel:poly}
\stackelpoly is \sigmaSecond-complete.
\end{theorem}
 \iftr{\begin{proof}
\underline{Membership:} Membership in $\Sigma^P_2$ can be shown by providing a
polynomial time alternating Turing Machine, which switches only once from existential to
universal nodes during each run. Using existential nodes, we guess a leader plan
$\pi^L$ with cost of at most $c^L$, execute it (if possible), to reach a state
$s^L = I\apply{\lplan}$. As argued above, $|\pi^L|$ is polynomially bounded, so
$s^L$ can be computed in polynomial time. Once $s^L$ is computed, we switch to
universal nodes and then guess a follower plan $\pi^F$ of cost at most $c^F$
which is again at most polynomially long. We then determine whether $\pi^F$ is
applicable in $s^L$ and whether $s^L\apply{\fplan} \subseteq G$.
If so we return false, otherwise true.

\underline{Hardness:} We reduce from the corresponding restricted QBF problem --
which is to determine whether formulae of the form $\exists x_i \forall y_j
\phi$ are satisfiable. W.l.o.g.\ we can assume that $\phi$ is in
DNF.\footnote{Satisfiability of $\exists x_i \forall y_j \phi$ is trivial if $\phi$ is in CNFs as tautology is trivial for CNFs.} 
Let $\psi_i$ be the $i$th cube of $\phi$.
We construct a Stackelberg task $\stask = \tuple{\facts, \lactions, \factions,
\init, \fgoal}$, in which the leader selects the $x_i$ variable
assignment, and the follower tries to find a $y_j$ assignment making $\phi$
evaluate to false:
\begin{align*}
\facts = &\{T^x_i, F^x_i, S^x_i \mid x_i\} \cup \{T^y_j, F^y_j, S^y_j \mid y_j\} \cup \{c_i \mid \psi_i \in \phi\}
\end{align*}
The initial state is $ \init = \{\}$.
The leader actions consists of:
\begin{itemize}
\item $sel^x_i\var{-T}$ with $pre(\var{sel^x_i-T}) = \{\neg S^x_i\}$ and \\ $add(\var{sel^x_i-T}) = \{S^x_i,T^x_i\}$
\item $sel^x_i\var{-F}$ with $pre(\var{sel^x_i-F}) = \{\neg S^x_i\}$ and \\ $add(\var{sel^x_i-F}) = \{S^x_i,F^x_i\}$
\end{itemize}
The follower has the following actions
\begin{itemize}
\item $sel^y_j\var{-T}$ with $pre(\var{sel^y_j-T}) = \{\neg S^y_j\}$ and \\ $add(\var{sel^y_j-T}) = \{S^y_j,T^y_j\}$
\item $sel^y_j\var{-F}$ with $pre(\var{sel^y_j-F}) = \{\neg S^y_j\}$ and \\ $add(\var{sel^y_j-F}) = \{S^y_j,F^y_j\}$

\item $val_{c_i}^{j}$ with $add(val_{c_i}^{j}) = \{c_i\}$, where
$l_j$ is the $j$-th literal in the $i$-th cube.
\begin{itemize}
\item If it is positive literal then
$pre(val_{c_i}^{j}) = \{F^l_j\}$
\item If it is a negative literal, then
$pre(val_{c_i}^{j}) = \{T^l_j\}$ 

\end{itemize}

\item $valS_{c_i}^{j}$ with $add(valS_{c_i}^{j}) = \{c_i\}$ and
$pre(valS_{c_i}^{j}) = \{\neg S^x_k\}$, where $l_j$ is the $j$-th literal in the
$i$-th cube, and $l_j \in \{ x_k, \neg x_k\}$ for some $k$.

\end{itemize}
All actions have cost $1$.
We set the goal to $G = \{c_i \mid \text{for every cube } i \text{ in }\phi\}$.
We lastly set $\lbound = |\{x_i | i\}|$ and $\fbound = |\{y_j \mid j\}| + \#cubes + 1$.

The leader chooses the $x_i$ assignment by executing either $sel^x_i\var{-T}$ or
$sel^x_i\var{-F}$ for every $x_i$ variable. After that, the follower can select
truth values of the $y_j$ variables using the $sel^y_j\var{-T}$ and
$sel^y_j\var{-F}$ actions, in attempts to make one of the $val_{c_i}$ actions
for every cube $c_i$ applicable. If this is possible, the respective cubes must
be violated. If all cubes evaluate to false, then so does the overall formula
$\phi$. The additional $valS_{c_i}^{j}$ actions are necessary to force the
leader to choose an assignment to all $x_i$ variables. Otherwise, unassigned
$x_i$ variables could make it impossible for the follower to find violations to
all cubes.
The value of $\lbound$ allows the leader to choose an assignment for all $x_i$
variables.
If the follower can reach her goal, she obviously has a plan with cost less than
$\fbound$.
If there is a leader plan $\lplan$ where $\fcost(\lplan) \geq \fbound$, then the
formula $\exists x_i \forall y_j \phi$ is satisfiable.
\end{proof}

 }

This result implies that, unless $\np = \sigmaSecond$ which would collapse the
polynomial hierarchy~\cite[Theorem~5.6]{Arora2007Complex}, polynomial
compilations of Stackelberg planning into classical planning come with a
worst-case exponential plan-length increase.

\subsection{Bylander's Syntactic Restrictions}

\citeauthor{Bylander1994}~(\citeyear{Bylander1994}) studied the complexity of
classical planning under various syntactic restrictions, drawing a concise
borderline between planning's tractability and infeasibility. 
\citeauthor{Bylander1994} distinguishes between different planning task classes
based on the number of action preconditions and effects, and the
existence of negative preconditions or effects. Table~\ref{table:overview}
provides an overview of the main classes.
Here, we take up his analysis and show that even for the classes where classical
planning is tractable, Stackelberg may not be. We consider \stackelsat and
\stackelopt in this order.
\begin{definition}
Let $m, n \in \naturals_0 \cup \{\infty\}$. $\stackelsat^m_n$ is the problem of
deciding $\stackelsat$ for Stackelberg tasks so that $|\pre(a)| \leq m$ and
$|\add(a)| + |\del(a)| \leq n$ hold for all actions $a$. If $m$ is preceded by
``$+$'', actions have no negative preconditions. If
$n$ is preceded by ``$+$'', actions have no delete effects. $\stackelopt^m_n$ is defined similarly.
\end{definition}
We omit $m$ ($n$) if $m = \infty$ ($n = \infty$).
We consider only cases where the classical-planning decision problems are in
\np.
Stackelberg planning is \pspace-hard when classical planning is.

\subsubsection{Plan Existence}\hfill\\
\citeauthor{Bylander1994}~(\citeyear{Bylander1994}) has shown that $\plansat$ is
already \np-complete for tasks with actions that even have just a single
precondition and a single effect. Here we show that the corresponding
Stackelberg decision problem is even one step above in the polynomial
hierarchy:

\begin{theorem}
$\stackelsat^{1}_{+1}$ is \sigmaSecond-complete.
\label{thm:stackel-sat-1-1}
\end{theorem}

 \iftr{
\begin{proof}
\underline{Membership:} As there are no delete effects, no action ever needs to
be applied more than once. Hence, if a leader plan satisfying
$\stackelsat^{1}_{+1}$ exists, then there exists one whose size is polynomially
bounded.
The same also holds for the follower.
To decide $\stackelsat^{1}_{+1}$, we can thus use a similar approach as
in Theorem~\ref{stackel:poly}.

\underline{Hardness:} We show hardness again via a reduction from the
satisfiability of restricted QBF of the form $\exists x_i \forall y_j \phi$,
assuming $\phi$ to be in DNF. Similar to the proof of
Theorem~\ref{stackel:poly}, the idea of our construction is to let the leader
choose an assignment to $x_i$, which the follower needs to counter by finding an
assignment to $y_j$ that makes $\phi$ false. 

The Stackelberg problem is defined as follows: The state variables are $\facts =
\{T_i^x, F_i^x, T_j^y, F_j^y, C_k\}_{i,j,k}$ for appropriately ranging $i, j, k$. The initial
state is $\init = \emptyset$. The follower's goal is $\fgoal = \{C_k \mid \text{for each
cube } k \text{ in } \phi\}$. The leader can choose the truth value for each
$x_i$:
via either $sel^x_i\var{-}T$ with $\pre(sel^x_i\var{-}T) = \{\neg F_i^x\}$ and $\add(sel^x_i\var{-}T) = \{T_i^x\}$ or 
$sel^x_i\var{-}F$ with $\pre(sel^x_i\var{-}F) = \{\neg T_i^x\}$ and $\add(sel^x_i\var{-}F) = \{F_i^x\}$.
The follower can choose the truth value for each $y_j$ via either
$sel^y_j\var{-}T$ with $\pre(sel^y_j\var{-}T) = \{\neg F_j^y\}$ and $\add(sel^y_j\var{-}T) = \{T_j^y\}$ or 
$sel^y_j\var{-}F$ with $\pre(sel^y_j\var{-}F) = \{\neg T_j^y\}$ and $\add(sel^y_j\var{-}F) = \{F_j^y\}$, 
and she can make false each cube $c_k$ in $\phi$ via each literal $l_i \in c_k$
by $val_{c_k}^i$ where $\add(val_{c_k}^i) = \{c_k\}$ and
  \[
    \pre(val_{c_k}^i) = \begin{cases}
      \{\neg T^x_i\} & \text{if } l_i = x_i \\
      \{\neg F^x_i\} & \text{if } l_i = \neg x_i \\
      \{F^y_i\} & \text{if } l_i = y_i \\
      \{T^y_i\} & \text{if } l_i = \neg y_i 
    \end{cases}
  \]
This task obviously satisfies the $\stackelsat^1_{+1}$ planning task
restrictions. Moreover, note that $\exists x_i \forall y_j \phi$ is
satisfiable iff the answer to $\stackelsat^1_{+1}$ is yes.
\end{proof}

 }

\citeauthor{Bylander1994}~(\citeyear{Bylander1994}) has shown that $\plansat$ is
polynomial if only positive preconditions and only a single effect per action are
allowed. Even under these restrictive conditions, $\stackelsat$ however still
remains intractable:

\begin{theorem}
$\stackelsat^{+}_{1}$ is \np-complete.
\label{thm:stackel-sat-+-1}
\end{theorem}
 \iftr{\begin{proof}
\underline{Membership:} Due to the restrictions, no action needs to be executed more than once.
Hence, as before, the consideration of polynomially length-bounded plans
suffices for answering Stackelberg plan existence for this class of tasks. To
solve $\stackelsat^{+}_{1}$, non-deterministically choose a (polynomially
bounded) leader plan $\lplan$ and construct the corresponding follower task
$\ftask{\lplan}$. This can be done in polynomial time. \plansat for
$\ftask{\lplan}$ can be answered in (deterministic) polynomial time
\cite{Bylander1994}. Return true if the follower task is unsolvable, otherwise
return false.

\underline{Hardness:} By reduction from Boolean satisfiability. Let $\phi$ be a
CNF over propositional variables $x_1, \dots, x_n$. We construct a Stackelberg
task, in which the leader decides the variable assignment, and the follower
evaluates the chosen assignment so that it has a plan iff the leader's chosen
assignment does not satisfy $\phi$. The task is composed of the state variables
$\facts = \{T_i, F_i \mid 1\leq i \leq n\} \cup \{U\}$. The initial state is $\init
= \{T_i, F_i \mid 1 \leq i \leq n\}$. The follower's goal is $\goal = \{U\}$. The
leader chooses the truth assignment by removing the unwanted value via either 
$sel_i\var{-}T$ with $\pre(sel_i\var{-}T) = \{T_i\}$ and $\del(sel_i\var{-}T) = \{F_i\}$
or
$sel_i\var{-}F$ with $\pre(sel_i\var{-}F) = \{F_i\}$ and $\del(sel_i\var{-}F) = \{T_i\}$.
The follower can evaluate each cube $C_k \in \phi$ via $val_k$ where
$\add(val_k) = \{U\}$ and $\pre(val_k) = \{ F_i \mid x_i \in C_k\} \cup \{ T_i \mid
\neg x_i \in C_k\}$ (the negation of the clause). 
The construction obviously fulfills the syntactic restrictions. Moreover, the
answer to $\stackelsat^{+}_{1}$ is yes iff $\phi$ is satisfiable.
\end{proof}

 }

Stackelberg plan-existence however becomes easy, when forbidding preconditions
throughout. While this class of tasks seems to be trivial at first glance,
optimal Stackelberg planning actually remains intractable as we show below.

\begin{theorem}
$\stackelsat^{0}$ is polynomial.
\label{thm:stackel-sat-0}
\end{theorem}
 \iftr{
\begin{proof}
Any $v \in V \setminus G$ can be ignored.
Consider the set $L^F$ of all follower actions $a^F \in \factions$ with $del(a^F) = \emptyset$.
The last action of any follower plan must be an action $a^F \in L^F$, i.e., if $L^F = \emptyset$, the follower can only use the empty plan.
Otherwise, the follower can always execute all $a^F \in L^F$ as its last actions.
We can thus remove any $v \in add(a^F)$ for any $a^F \in L^F$ from consideration (remove it from $\fgoal$ and $\facts$).
We can now recalculate $L^F$ and repeat this process until $L^F = \emptyset$.
This process terminates after polynomially many steps.
If at this point $\fgoal \not \subseteq \init$, the follower has no plan for the empty leader plan.
Otherwise, the follower has no plan iff there is an action $v \in \fgoal$ s.t.\ there is $a^L \in \lactions$ with $v \in del(a^L)$.
The leader plan is then $a^L$.
\end{proof}

 }

\subsubsection{Optimal Planning}\hfill\\
As per Proposition~\ref{prop:sat-opt-reducible}, optimal Stackelberg planning is in general
at least as hard as deciding plan existence. All intractability results shown
for $\stackelsat$ carry over to $\stackelopt$. As in all classes
analyzed in the previous section, the consideration of polynomially
length-bounded plans is sufficient for hardness, \sigmaSecond yields a sharp upper bound to
the complexity of $\stackelopt$, per Theorem~\ref{stackel:poly}.

\begin{corrolary}
$\stackelopt^1_{+1}$ is \sigmaSecond-complete.
\label{cor:stackel-opt-1-1}
\end{corrolary}

 \iftr{\begin{proof}
Follows directly from Theorem~\ref{thm:stackel-sat-1-1}.
\end{proof}

 }

The results for $\stackelsat$ only provide a lower bound to the
complexity of $\stackelopt$. This lower bound may be strict as demonstrated by
Thm.~\ref{thm:stackel-opt-+-1} and \ref{thm:stackel-opt-0-2}:

\begin{theorem}
$\stackelopt^{+1}_{1}$ is \sigmaSecond-complete.
\label{thm:stackel-opt-+-1}
\end{theorem}
 \iftr{
\begin{proof}
\underline{Membership:} As argued in Theorem~\ref{thm:stackel-sat-+-1}, the
consideration of polynomially long plans suffices to answer
$\stackelopt^{+1}_{1}$. Membership then follows via the
procedure sketched in Theorem~\ref{stackel:poly}.

\underline{Hardness:} Reduction from the satisfiability problem for restricted
QBFs $\exists x_i \forall y_j \phi$, assuming $\phi$ to be in DNF.
Let $n$ be the number of $x_i$ variables and $m$ the number of $y_j$ variables.
For convenience of notation, we assume for this proof (and only this proof) that the $y_j$ variables are numbered from $y_{n+1}$ to $y_{n+m}$.
Let $k$ be the number of cubes in $\phi$. The idea of our
Stackelberg planning task construction is similar to all prior proofs. The state
variables are $\facts = \{ T_i, F_i \mid 1 \leq i \leq n+m \} \cup \{ S_{n+i} \mid 1
\leq i \leq m\} \cup \{ C_j \mid 1 \leq j \leq k\}$. The initial state is $\init =
\{ T_i, F_i \mid 1 \leq i \leq n\}$. The follower's goal is $\fgoal = \{ S_{n+i} \mid
1 \leq i \leq m\} \cup \{ C_i \mid 1 \leq i \leq k\}$. The leader can choose the
$x_i$ truth assignments by removing the unwanted value ($1 \leq i \leq n$) via 
$sel_i\var{-}T$ with $\pre(sel_i\var{-}T) = \{T_i\}$ and $\del(sel_i\var{-}T) = \{F_i\}$ and 
$sel_i\var{-}F$ with $\pre(sel_i\var{-}F) = \{F_i\}$ and $\del(sel_i\var{-}F) = \{T_i\}$.
The follower can choose the truth value for each $y_j$ ($n+1 \leq i \leq n+m$) via
$sel_{i}\var{-}T$ with $\add(sel_{i}\var{-}T) = \{T_{i}\}$ or
$sel_{i}\var{-}F$ with $\add(sel_{i}\var{-}F) = \{F_{i}\}$.
The follower can indicate that $y_j$ has been assigned through ($n+1 \leq i \leq n + m$): via
$done_{i}\var{-}T$ with $\pre(done_{i}\var{-}T) = \{T_i\}$ and $\add(done_{i}\var{-}T) = \{S_{i}\}$ or
$done_{i}\var{-}F$ with $\pre(done_{i}\var{-}F) = \{F_i\}$ and $\add(done_{i}\var{-}F) = \{S_{i}\}$, 
and, finally, it can evaluate each cube $c_j$ in $\phi$ through each of the literals $l_i \in c_k$
by $val_{j}^i$ where $\add(val_{j}^i) = \{C_j\}$ and
if $l_i$ is positive, then $\pre(val_{j}^i) = \{F_i\}$
and otherwise if $l_i$ is negative, then $\pre(val_{j}^i) = \{T_i\}$.
All actions have unit cost.
Note that the construction satisfies the syntactic restrictions of
$\stackelopt^{+1}_{1}$.
In order to reach its goal, the follower must execute one of the $done_i$
actions for each variable $y_j$, which in turn requires executing one of the
$sel_i$ actions for each variable $y_j$, and it must execute one of the $val_j$
actions for each cube. Hence, there is no follower plan shorter than $2m + k$.
Plans which assign some $y_j$ variable multiple values are possible, but they
have to be longer than $2m + k$. If the follower has a plan with exactly that
length, then the formula $\phi$ can be falsified given the $x_i$ assignments
chosen by the leader.
So, let $\fbound = 2m + k + 1$ and $\lbound = n$. The latter suffices to allow
the leader to choose an assignment for every $x_i$. The answer to
$\stackelopt^{+1}_{1}$ for these bounds is yes iff the QBF is satisfiable.
\end{proof}

 }

\begin{theorem}
$\stackelopt^{0}_{2}$ is \sigmaSecond-complete.
\label{thm:stackel-opt-0-2}
\end{theorem}
 \iftr{\begin{proof}
\underline{Membership:} Since actions have no preconditions, it never makes
sense to execute an action more than once. As such, if a plan exists, a
polynomially long plan exists as well. We can thus use the same algorithm as in
Theorem~\ref{stackel:poly}.

\underline{Hardness:} We again reduce from satisfiability of QBF formulae of the
form $\exists x_i \forall y_j \phi$. We assume
that $\phi$ is in DNF. We further assume that the variables $x_i$ are numbered $1$ to $n$ and the $y_j$ are numbered $n+1$ to $n+m$.

Let $k$ be the total number of cubes in $\phi$. Our
Stackelberg task encoding follows once again also the same idea as before. The
state variables are $\facts = \{notT^x_i, notF^x_i, S^x_i \mid 1 \leq i \leq n\}
\cup \{notT^y_j, notF^y_j, S^y_j \mid n+1 \leq i \leq n+m\} \cup \{C_i \mid 1
\leq i \leq k \}$. The initial state is $\{notT^x_i, notF^x_i \mid 1 \leq i \leq
n\} \cup \{notT^y_j, notF^y_j \mid n+1 \leq i \leq n+m\} $. The follower's goal
is $\fgoal = \{ notT^x_i, notF^x_i, S^x_i \mid 1\leq i \leq n \}  \cup \{ notT^y_j,
notF^y_j, S^y_j \mid n+1\leq i \leq n+m \} \cup \{ C_j \mid 1 \leq j \leq k\}$. We
then add the following leader actions
$sel_i\var{-T}$ with $add(\var{sel_i-T}) = \{notF_i\}$ and $del(\var{sel_i-T}) = \{notT_i\}$ and 
$sel_i\var{-F}$ with $del(\var{sel_i-F}) = \{notT_i\}$ and $del(\var{sel_i-F}) = \{notF_i\}$.
For the follower, we add the following actions: 
(1) to assume the truth value of a variable ($x_i$ or $y_j$) to be $B \in \{T,
F\}$ ($1 \leq i \leq n+m$): $assume_i\var{-B}$ with $add(\var{assume_i-B}) =
\{S_i\}$ and $del(\var{assume_i-B})=\{notB_i\}$,
(2) to evaluate the $i$-th cube to false by using the assumption that literal
$l_j \in C_i$ is false:
$add(val_{C_i}^{j}) = \{C_i\}$ and 
if $l_j$ is a positive literal, then
$del(val_{C_i}^{j}) = \{notT_j\}$ and 
otherwise if it is a negative literal, then
$del(val_{C_i}^{j}) = \{notF_j\}$.
Note that if the assumption is indeed
satisfied, the delete effect becomes a noop.
(3) And finally, to revert an assumption: $revert_i\var{-B}$ with
$add(\var{revert_i-B}) = \{notB_i\}$
All actions have cost $1$.

To reach the goal, the follower needs to perform three things:
(1) Make an assumption about the value of every $x_i$ and $y_j$ variable.
(2) Evaluate all cubes to false by picking one literal and forcing its negation to be true.
(3) Unassign every variable by applying revert according to the deleted facts.
All in all, each follower plan must contain at least $2(n+m) + k$ actions. If
there is a plan with exactly this length, then all the chosen $val_j$ actions
had to use an already assumed variable-truth-value; and every variable must have
exactly one assumed truth value; in particular, the follower plan must assume the truth
value of the $x_i$ variables that was chosen by the leader. Hence, each such
plan corresponds to a violating assignment to $\phi$. If, on the other hand, for
the $x_i$ assignment chosen by the leader $\forall y_j: \phi$ is true, the
length of an optimal follower plan must exceed $2(n+m) + k$, as making false all
cubes in $\phi$ then requires assuming both truth-values for at least one
variable (meaning additional 2 actions).
The answer to $\stackelopt_{2}^0$ for $\lbound = n$ and $\fbound
= 2(n+m)+k+1$ is yes iff the QBF is satisfiable. 
\end{proof}

 }

Optimal Stackelberg planning remains intractable even when all actions have no
preconditions and may have only at most one effect.

\begin{theorem} $\stackelopt^{0}_{1}$ is \np-complete in general, but polynomial
when additionally assuming unit cost.
\label{thm:stackel-opt-0-1}
\end{theorem}
 \iftr{\begin{proof} For the leader it only makes sense to execute actions with a
deleting effect and for the follower actions with an adding effect. 
More specifically, let $\goal' := \goal \cap \init$. In order to increase the
plan cost of the follower, the leader needs to apply actions that delete some
fact from $\goal'$. On the other hand, the follower has to apply an action for
every $\goal \setminus \goal'$, and in addition an action for every fact from
$\goal'$ the leader has deleted. If all costs are equal, the leader either has
to delete a state variable that the follower cannot add or the cost bound
$\lbound$ and the available actions must allow to delete at least $\fbound +
|\goal'| - |\goal|$ many facts from $\goal'$. Otherwise the leader cannot solve
the task. This can be checked in polynomial time.
Suppose that actions may have non-unit cost.

\underline{Membership:} We can non-deterministically guess a subset of the
leader actions of cost at most $\lbound$ and execute them. From the resulting
state $s$, the follower has to execute her actions that make the state variables
in $G \setminus s$ true. We can select per variable the cheapest action and add
the costs up. We return true if this is above $\fbound$.

\underline{Hardness:} We reduce from integer knapsack~\cite[MP10]{GareyBook1979}. Let $U = \{u_1, \dots,u_n\}$ be a set of objects, $s: U \mapsto \mathbb N^+$ be their sizes, $v: U \mapsto \mathbb N^+$ their values, $B$ the size limit, and $K$ the minimal
desired total value. We construct a Stackelberg task following the same intuition
as in the proof of Theorem~\ref{thm:stackel-sat-+-1}: the leader picks a
possible solution and the follower's plans correspond to the evaluation of this
solution. We set facts $\facts$, initial state $\init$, and goal $\fgoal$ all to be the set of objects $U$, i.e., $\facts = \init = \fgoal = U$. The leader has for every $u_i$ an
action $sel_{u_i}$ with $del(sel_{u_i}) = \{u_i\}$ and cost $s(u_i)$. The
follower has for every $u_i$ an action $take_{u_i}$ with $add(take_{u_i}) =
\{u_i\}$ and cost $v(u_i)$. 
We set $\lbound = B$ and $\fbound = K$. 
The leader's selection of $sel_{u_i}$ actions encodes a set of objects $S
\subseteq U$ fitting the size limit, i.e., $\sum_{u \in S} s(u) \leq B$.
In order to achieve its goal, the follower needs to take (at least) all the
objects selected by the leader, resulting in a cost of at least $\sum_{u \in S}
v(u)$. Therefore, the leader selection is a solution to the bin-packing problem
if the follower's optimal plan cost is at least $K = \fbound$. The answer to $\stackelopt^0_{1}$ is yes iff the bin-packing instance has a
solution.
\end{proof}

 }

 \renewcommand{\lbound}{B^P}
\renewcommand{\fbound}{B^M}
\section{Complexity of Meta Operator Verification}

\citeauthor{Pham2023MetaOperators}~(\citeyear{Pham2023MetaOperators}) have
recently leveraged Stackelberg planning for synthesizing \emph{meta-operators}
in classical planning. Meta-operators, like macro-actions
\cite{Fikes1971STRIPS}, are artificial actions that aggregate the effect of
action sequences, therewith introducing shortcuts in state-space search.
Formally, given a classical planning task $\task$ and an action $\meta$ that is
not in $\task$'s action set, $\meta$ is a \defined{meta-operator
for $\task$} if, for every state $s \models \pre(\meta)$ that is reachable from
$\init$, there exists a sequence $\plan$ of $\task$'s actions such that
$s\apply{\meta} = s\apply{\pi}$. Whether a given $\meta$ is a meta-operator can
be \emph{verified} by solving a Stackelberg planning task.

Here, we consider the question whether using an expressive and computationally
difficult formalism like Stackelberg planning is actually necessary. For this,
we determine the computational complexity of meta-operator synthesis and compare
it to that of Stackelberg planning.

\begin{definition}[Meta-Operator Verification]\label{base:meta}
Given $\task$ and a fresh action $\meta$. \metaverif is the problem of
deciding whether $\meta$ is a meta-operator for $\task$.
\end{definition}

Like for Stackelberg planning, the complexity of meta-operator verification in
general remains the same as that of classical planning:

\begin{theorem}
\metaverif is \pspace-complete.
\label{thm:metaverif-base}
\end{theorem}
 \iftr{

\begin{proof}

\underline{Membership:} Iterate over all states in $\task$ (which only requires
to store the currently considered state, i.e., can be done in polynomial space).
For each state $s$: (1) check if $s \models \pre(\meta)$, and if so (2) check
whether $s$ is reachable from $\init$, and if this is also the case, (3) check
whether $s\apply{\meta}$ is reachable from $s$. (1) can be clearly tested in
polynomial space. (2) and (3) can be done in polynomial space with a small
modification of the algorithm used to show plan existence in classical planning:
instead of using the subset-based goal termination test, we enforce equality,
terminating only at states $t$ with (2) $t = s$ respectively (3) $t =
s\apply{\meta}$. We return true if (3) was satisfied for states tested, and
false otherwise.

\underline{Hardness:} We reduce from \plansat. Let $\task = \tuple{\facts,
\actions, \init, \goal}$ be a classical planning task. Let $g$ be a fresh state
variable, and $a_g$ be a fresh action. We create a new planning task $\task' =
\tuple{\facts \cup \{g\}, \actions \cup \{a_g\}, \init, \{g\}}$ where $\pre(a_g)
= \goal$, $\add(a_g) = \{g\}$, $\del(a_g) = \facts$. Note that $\task$ is
solvable iff $\task'$ is solvable. We define a new meta-operator $\meta$ for
$\task'$, setting $\pre(\meta) = \{ p | p \in \init \} \cup \{ \neg p | p \in
\facts \setminus \init \}$, $\add(\meta) = \{g\}$, and $\del(\meta) = \facts$.
Obviously, $\meta$ is a meta-operator for $\task'$ iff $\task'$ is solvable,
what shows the claim.

\end{proof}

 }

In other words, meta-operator verification could as well be compiled directly
into a classical rather than a Stackelberg planning task. But how difficult or
effective would such a compilation be? To shed light on this question, we again
turn to a length bounded version of the problem.

\begin{definition}[Polynomial Meta-Operator Verification]
Given $\task$ with non-$0$ action costs, a fresh action $\meta$, and two
binary-encoded numbers $\lbound, \fbound \in \mathbb N_0$ that are bounded by
some polynomial $p \in \mathcal O(\ell^k)$ for $\ell = |\facts| + |\actions|$.
\polymetaverif is the problem of deciding whether for all states $s \models
\pre(\meta)$ reachable from $\init$ with a cost of at most $\lbound$, there exists
$\pi$ with $c(\pi) \leq \fbound$ and $s\apply{\pi} =
s\apply{\meta}$.
\end{definition}
The parameters $\lbound$ and $\fbound$ define the perimeter around the initial
state respectively the reached state under which the meta-operator conditions
are to be verified. As for Stackelberg planning, we require that the cost of all
actions is strictly positive, which together with the cost bounds ensures that the
radius of the perimeter is polynomially bounded.

Polynomial meta-operator verification too is on the second level of the
polynomial hierarchy. Again, this means that under the assumption that the
polynomial hierarchy does not collapse, polynomial compilations of meta-operator
verification into classical planning in the worst case, come with an exponential
plan-length blow-up.

\begin{theorem}
\polymetaverif is $\Pi^P_2$-complete.
\label{thm:metaverif-poly}
\end{theorem}
 \iftr{
\begin{proof}
\underline{Membership:} Membership in $\Pi^P_2$ can be show by providing an alternating Turing Machine, which switches only once from universal to existential nodes during each run.
Using universal nodes, we guess a plan of cost at most $\lbound$, execute it (if possible), to reach a state $s^P$ and check whether $s^P \models pre(\meta)$.
If not, return true (as we can not disprove validity with this trace).
If $s^P \models pre(\meta)$, then using existentially quantified decision nodes, guess a plan of cost at most $\fbound$, check its applicabiltiy (else return false) and whether it reaches $s^P[[\meta]]$.
If so, return true, else false.

\underline{Hardness:} We reduce from the respective restricted QBF satisfiability problem -- which are formulae of the form $\forall x_i \exists y_j \phi$.
We can assume that $\phi$ is in 3-CNF.
We define the state variables
\begin{align*}
V = &\{B\} \cup \{T^x_i, F^x_i, S^x_i \mid x_i\} \cup \{T^y_j, F^y_j, S^y_j \mid y_j\} \\
{} &\cup \{cl_i \mid \text{for every clause } i \text{ in }\phi\}
\end{align*}
The initial state is $\{B\}$.
We then define actions 
\begin{itemize}
\item $sel^x_i\var{-T}$ with $pre(\var{sel^x_i-T}) = \{\neg S^x_i, B\}$ and \\ $add(\var{sel^x_i-T}) = \{S^x_i,T^x_i\}$
\item $sel^x_i\var{-F}$ with $pre(\var{sel^x_i-F}) = \{\neg S^x_i, B\}$ and \\ $add(\var{sel^x_i-F}) = \{S^x_i,F^x_i\}$
\item $\var{do-block}$ with $pre(\var{do-block}) = \{B\} \cup \{S^x_i \mid x_i\}$ and \\ $del(\var{do-block}) = \{B\}$
\item $sel^y_j\var{-T}$ with $pre(\var{sel^y_j-T}) = \{\neg S^y_j,\neg B\}$ and \\ $add(\var{sel^y_j-T}) = \{S^y_j,T^y_j\}$
\item $sel^y_j\var{-F}$ with $pre(\var{sel^y_j-F}) = \{\neg S^y_j, \neg B\}$ and \\ $add(\var{sel^y_j-F}) = \{S^y_j,F^y_j\}$
\item $val_{cl_i}^{j}$ with $add(val_{cl_i}^{j}) = \{cl_i\}$.
Let $l_j$ be the $j$-th literal in the clause $i$.
\begin{itemize}
\item If it is a positive literal then
$pre(val_{cl_i}^{j}) = \{\neg B, T^l_j\}$
\item If it is a negative literal, then
$pre(val_{cl_i}^{j}) = \{\neg B, F^l_j\}$ 
\end{itemize}
\item $\var{re-block}$ with $pre(\var{re-block}) = \{\neg B\} \cup \{S^y_j \mid y_j\}$, \\
$add(\var{re-block}) = \{B\}$, and \\ $del (\var{re-block}) = \{T^y_j, F^y_j \mid y_j\}$
\end{itemize}
All actions have cost $1$.

We then ask whether the meta operator $\meta$ with
$pre(\meta) = \{B\} \cup \{S^x_i \mid x_i\} \cup \{\neg S^y_j \mid y_j\}$ and \linebreak
$add(\meta) = \{cl_i \mid \text{for every clause } i \text{ in } \phi\} \cup \{S^y_j \mid y_j\}$
is valid under the cost limits $\lbound = |\{x_i \mid x_i\}|$ and \linebreak $\fbound = |\{y_j \mid y_j\}| + |\{i \mid \text{for every clause } i \text{ in } \phi\}| + 2$

We claim that the meta operator $\meta$ is valid if and only if the formula $\phi$ is satisfiable.
To validate $\meta$, we have to consider any reachable state $s^P$ (with cost at most $\lbound$) in which $B$, all the $S^x_i$, but none of the $S^y_j$ are true. Since the block variable $B$ has to be true in this state, we cannot have executed $\var{do-block}$ -- otherwise we would also require a $\var{re-block}$ which exceeds together with the necessary $sel^x$ actions the cost limit $\lbound$.
Thus in any such state $s^P$, a truth value for all the $x_i$ variables must have been selected, but for none of the $y_j$ variables.

For $\meta$ to be valid, for any such $s^P$, we have to find a plan that reaches $s^P[[\meta]]$.
Given the effects of $\meta$, this means that we have to select a value for all $y_j$ variables and to satisfy all clauses (via the $cl_i$ variables).
As the first action of any such plan, we have to perform $\var{do-block}$ -- as all other actions (except the $sel^x$ which we cannot execute anymore at this point) require $\neg B$.
We then have to select truth values for the variables $y_j$ using the $sel^y$ actions.
At this point a single, non-modifiable valuation of the $x_i$ and $y_j$ has been chosen. If that valuation indeed satisfies $\phi$, then there is an 
appropriate selection of $val_{cl_i}$ actions that marks all clauses as satisfied.
Lastly, the plan has use the $\var{re-block}$ action to clear the information on how we set the truth values for the $y_j$ variables and to make the variable $B$ true again.
This is required as we have to reach $s^P[[\meta]]$ exactly.
In essence, the $\var{re-block}$ action allows us not to ``leak'' any information on how we selected the truth values of the $y_j$ variables out of the execution of the meta operator.

  Since every valuation of the $x_i$ variables corresponds to a reachable state (within the $\lbound$ cost perimeter), $\meta$ is valid iff for every such valuation we can find a plan that sets the $y_j$ in a way that satisfies all clauses in the formula.
If $\meta$ is not valid, we can on the other hand find a valuation of the $x_i$ for which we cannot achieve the target state of $\meta$, thus it is impossible to set the $y_j$ to satisfy the formula.
Thus, $\meta$ is valid if and only if the QBF is true.
\end{proof}
 }

Note that \polymetaverif is therefore in the co-complexity-class of polynomial
Stackelberg plan-existence, i.e., they belong to co-classes on the same level of
the polynomial hierarchy. This may not be surprising given the subtle difference between meta-operator
verification and Stackelberg plan existence: while
the latter asks for the existence of a (leader) action sequence where all
induced (follower) action sequences satisfy some property, meta-operator
verification swaps the quantifiers.

We want to point out that this duality can be exploited further, showing
analogous results for \citeauthor{Bylander1994}'s~(\citeyear{Bylander1994}) task
classes. Contrary to Stackelberg planning, however, the identification of
tractable fragments is less useful for meta-operator verification due to the
lack of the monotonicity invariance of the meta-operator condition. An action
being a meta-operator in a task abstraction does not imply that the action is a
meta-operator in the original task, and vice versa. We hence do not further
explore this analysis here.

 \section{Related Work}

Stackelberg planning is strongly linked to conditional planning under partial
observability \cite{BonetG00} and non-deterministic actions
\cite{CimattiRT98}. 
Conditional plan existence in partially observable deterministic (POD) planning
can be seen as the co-problem to \stackelsat for the Stackelberg task where the
leader enumerates the initial states, and the follower has the POD task's
actions and goal.
As hinted at in the introduction, there is a simple reduction of Stackelberg
plan existence to FOND plan existence. In a nutshell, the
FOND planning task inherits the leader actions and contains a single action with
a non-deterministic effect per original follower action, and which
non-deterministically sets a termination flag $\mathtt{T}$ (initially false).
Leader planning and follower planning are split into two separate phases,
distinguished by auxiliary facts. In the leader phase, it is possible to apply
only leader actions and an action that will cause a transition into the follower
phase. In the follower phase, only the follower's non-deterministic action can
be applied. There is a Stackelberg  plan iff there is a strong cyclic plan for the
goal $\neg \fgoal \land \mathtt{T}$.

Both planning under partial observability and planning under non-deterministic
effects are more challenging problems than Stackelberg planning in the general
case, e.g., FOND planning is EXP-complete \cite{Littman97} and POND planning
even 2-EXP-complete \cite{rintanen2004complexity}.
But, there are interesting relations to our results under a polynomial plan
length bound.
\citeauthor{rintanen1999constructing} (\citeyear{rintanen1999constructing})
showed that polynomially-length-bounded conditional planning is $\piSecond$
complete, the co-result to our Thm.~\ref{stackel:poly}. His hardness proof is
very similar to ours, with technical differences owed to the different planning
formalisms. \citeauthor{bonet2010conformant}~(\citeyear{bonet2010conformant})
studied conditional planning with non-deterministic actions, proving that
polynomially bounded plan existence for conditional plans with at most $k$
branching points is $\sigmaP{2k+k}$-complete. Stackelberg planning corresponds
to $k=1$, the difference between determinism and non-determinism causing the
$\sigmaSecond$ vs. $\sigmaP{4}$ complexity results.
For conditional planning under partial observability,
\citeauthor{baral2000computational}~(\citeyear{baral2000computational}) showed
that plan existence is $\sigmaSecond$-complete.
\citeauthor{turner2002polynomial} (\citeyear{turner2002polynomial}) analyzed a
wide range of different planning formalisms under a polynomial plan-length
bound, but his formalism supported arbitrary boolean formulae as preconditions,
making even length-$1$ plan existence already \np-complete.

\section{Conclusion}

Stackelberg planning remains \pspace-complete, like classical planning, in
general, but is $\Sigma^P_2$ complete under a polynomial plan-length bound.
Hence, unless the polynomial hierarchy collapses to its first level, a
polynomial reduction of Stackelberg planning into classical planning is not
possible in general. We showed that Stackelberg planning remains intractable
under various syntactical restrictions, even in cases where classical planning
is known to be tractable. Lastly, we have proven similar results for
meta-operator verification, specifically \pspace-completeness in general and
$\Pi^P_2$-complete for the polynomial plan-length bounded case, with similar
implications as the results for Stackelberg planning.

\section*{Acknowledgments}

We thank the anonymous reviewers, whose comments helped to improve the paper.
This work has received funding from the European Union’s Horizon Europe Research
and Innovation program under the grant agreement TUPLES No 101070149. Marcel
Steinmetz would like to acknowledge the support of the Artificial and Natural
Intelligence Toulouse Institute (ANITI), funded by the French Investing for the
Future PIA3 program under the Grant agreement ANR-19-PI3A-000.

\bibliography{paper.bib}

\end{document}